\documentclass[10pt,twocolumn,letterpaper]{article}

\usepackage[pagenumbers]{iccv} % To force page numbers, e.g. for an arXiv version

\definecolor{iccvblue}{rgb}{0.21,0.49,0.74}
\usepackage[pagebackref,breaklinks,colorlinks,allcolors=iccvblue]{hyperref}

\def\paperID{8281} % *** Enter the Paper ID here
\def\confName{ICCV}
\def\confYear{2025}

\title{Dynamic Pyramid Network for Efficient Multimodal Large Language Model}
\author{Hao Ai\textsuperscript{1}, Kunyi Wang\textsuperscript{3}, Zezhou Wang\textsuperscript{3}, Hao Lu\textsuperscript{4}, Jin Tian\textsuperscript{3}, Yaxin Luo\textsuperscript{5} \\ Peng Xing\textsuperscript{6},
Jen-Yuan Huang\textsuperscript{7}, Huaxia Li\textsuperscript{8}, Gen Luo\textsuperscript{2}\\
\textsuperscript{1}Beihang University\quad \textsuperscript{2}Shanghai AI Laboratory\quad
\textsuperscript{3}KAUST \\
\textsuperscript{4}Hong Kong University of Science and Technology (Guangzhou)\ 
\textsuperscript{5}Technical University Of Denmark \\
\textsuperscript{6}Nanjing University of Science and Technology \quad
\textsuperscript{7}Peking University\quad
\textsuperscript{8}Xiaohongshu Inc
}

\begin{document}
\maketitle
\begin{abstract}
Multimodal large language models (MLLMs) have demonstrated impressive performance in various vision-language (VL) tasks, but their expensive computations still limit the real-world application.  To address this issue, recent efforts aim to compress the visual features to save the computational costs of MLLMs. However, direct visual compression methods, \emph{e.g.} efficient projectors, inevitably destroy the visual semantics in MLLM, especially in difficult samples.  To overcome this shortcoming, we propose a novel dynamic pyramid network (DPN) for efficient MLLMs. Specifically, DPN formulates MLLM as a hierarchical structure where visual features are gradually compressed with increasing depth. In this case, even with a high compression ratio, fine-grained visual information can still be perceived in shallow layers.  To maximize the benefit of DPN, we further propose an innovative Dynamic Pooling Experts (DPE) that can dynamically choose the optimal visual compression rate according to input features.  With this design, harder samples will be assigned larger computations, thus preserving the model performance.  To validate our approach, we conduct extensive experiments on two popular MLLMs and ten benchmarks. Experimental results show that DPN can save up to 56\% average FLOPs on LLaVA while further achieving +0.74\% performance gains. Besides, the generalization ability of DPN is also validated on the existing high-resolution MLLM called LLaVA-HR. The source code will be released at \url{https://github.com/aihao2000/DPN-LLaVA}.

\end{abstract}

% \url{https://anonymous.4open.science/r/DPN-LLaVA-8281}.    
\section{Introduction}
\label{sec:intro}

Multimodal large language models (MLLMs) have achieved significant progress in computer vision~\cite{lin2024vilapretrainingvisuallanguage,xu2024llavauhdlmmperceivingaspect,li2024llavaonevisioneasyvisualtask,liu2023visualinstructiontuning,luo2024feasteyesmixtureofresolutionadaptation,li2024minigeminiminingpotentialmultimodality,dai2023instructblipgeneralpurposevisionlanguagemodels,chu2023mobilevlmfaststrong,chu2024mobilevlmv2fasterstronger}, which continuously breakthroughs the performance limit of various vision-language (VL) tasks~\cite{goyal2017makingvvqamatter,hudson2019gqanewdatasetrealworld,lu2022learnexplainmultimodalreasoning,singh2019vqamodelsread,fu2024mmecomprehensiveevaluationbenchmark,liu2024mmbenchmultimodalmodelallaround,yu2023mmvetevaluatinglargemultimodal,yue2023mmmu,li2023seed,Li-hallucination-2023}. Among them, the mainstream paradigm is to insert  visual tokens from pre-trained encoder into the LLM, and then fine-tune the entire MLLM.   Despite effectiveness, expensive computation remains the main bottleneck of existing MLLMs in real-world applications~\cite{chen2024imageworth12tokens,xing2024pyramiddrop}, especially when image resolutions greatly increase~\cite{luo2024feasteyesmixtureofresolutionadaptation}.

To address this issue, increasing efforts are devoted to the efficient MLLM architectures~\cite{chen2024imageworth12tokens,lin2024moellavamixtureexpertslarge,chu2024mobilevlmv2fasterstronger,wu2024routingexpertslearningroute,dai2023instructblipgeneralpurposevisionlanguagemodels}.  Among them,  MoE-based approaches~\cite{lin2024moellavamixtureexpertslarge} aim to reduce the activated parameters of MLLMs via the mixture of expert, thereby improving the efficiency. However, these methods typically require expensive training stages for their MoE-based learning. Recently, another cheap and promising trend is to reduce the visual tokens of MLLMs via the efficient visual projector~\cite{chu2024mobilevlmv2fasterstronger,shang2024llavaprumergeadaptivetokenreduction,li2024tokenpackerefficientvisualprojector,bai2023qwenvlversatilevisionlanguagemodel}. For instance, Resampler~\cite{bai2023qwenvlversatilevisionlanguagemodel} compresses visual features into few learnable tokens via the cross-attentions.  By placing it before the LLM, the computation costs of MLLMs can be obviously saved due to the reduction of input sequence.

% existing approaches compress visual features into few tokens, thus obviously saving the computation costs of MLLMs.

% to save the computations. From this perspective, recent attempts often place an efficient visual projector~\cite{} before the LLM, and compress visual features into few tokens. In this way, 

\begin{figure*}[t]
  \centering
   \includegraphics[width=1.0\linewidth]{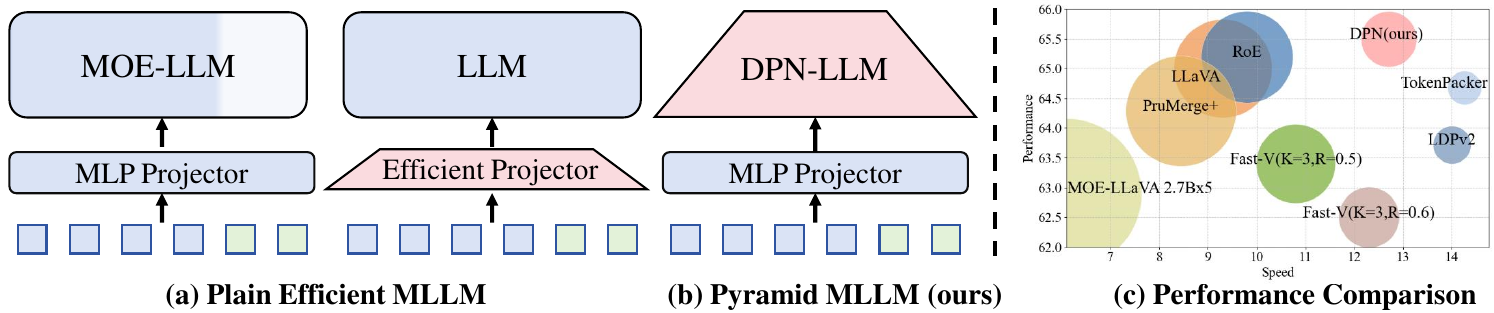}
   \caption{ \textbf{Illustration of existing plain MLLMs with our dynamic pyramid network (DPN).}  Compared to plain MLLMs, our DPN can  dynamically compress the input tokens in a hierarchical manner while greatly preventing the model performance.  }
   \label{fig1}
\end{figure*}

Despite the efficiency, we identify that  these methods are  sub-optimal and often sacrifice the visual representation. Specifically, most existing visual projectors are typically placed outside the LLM and
downsample visual features via their saliency~\cite{chu2024mobilevlmv2fasterstronger,li2024tokenpackerefficientvisualprojector}.  Such straightforward practice ignores the visual semantics and the textual prompt, thus easily destroying the fine-grained visual content, \emph{e.g.,} OCR words and small objects. More importantly, all samples are aligned with the same visual compression rate, \emph{e.g.,} 75\% for 2$\times$2 pooling, which inevitably hurts the understanding of hard samples and leads to the redundant computation on easy samples.

To overcome this shortcoming, we argue that the MLLM should be designed as a  pyramid structure, in which visual compression modules, \textit{e.g.,} pooling layers, are embedded into LLM in a hierarchical manner. In this way, visual information is mostly preserved in the shallow layers of LLM for fine-grained perception and gradually compressed in deeper layers for high-level understanding.  Compared to previous works~\cite{li2024tokenpackerefficientvisualprojector,chu2024mobilevlmv2fasterstronger,shang2024llavaprumergeadaptivetokenreduction,chen2024imageworth12tokens,wu2024routingexpertslearningroute}, even with a high compression rate,  MLLM can still capture critical visual content for vision-language understanding.  Besides, such hierarchical structure is also aligned with the most successful design principle in computer vision, \emph{e.g.,} Swin-Transformer~\cite{liu2021swintransformerhierarchicalvision}.

\begin{figure*}[ht]
  \centering
   \includegraphics[width=1.0\linewidth]{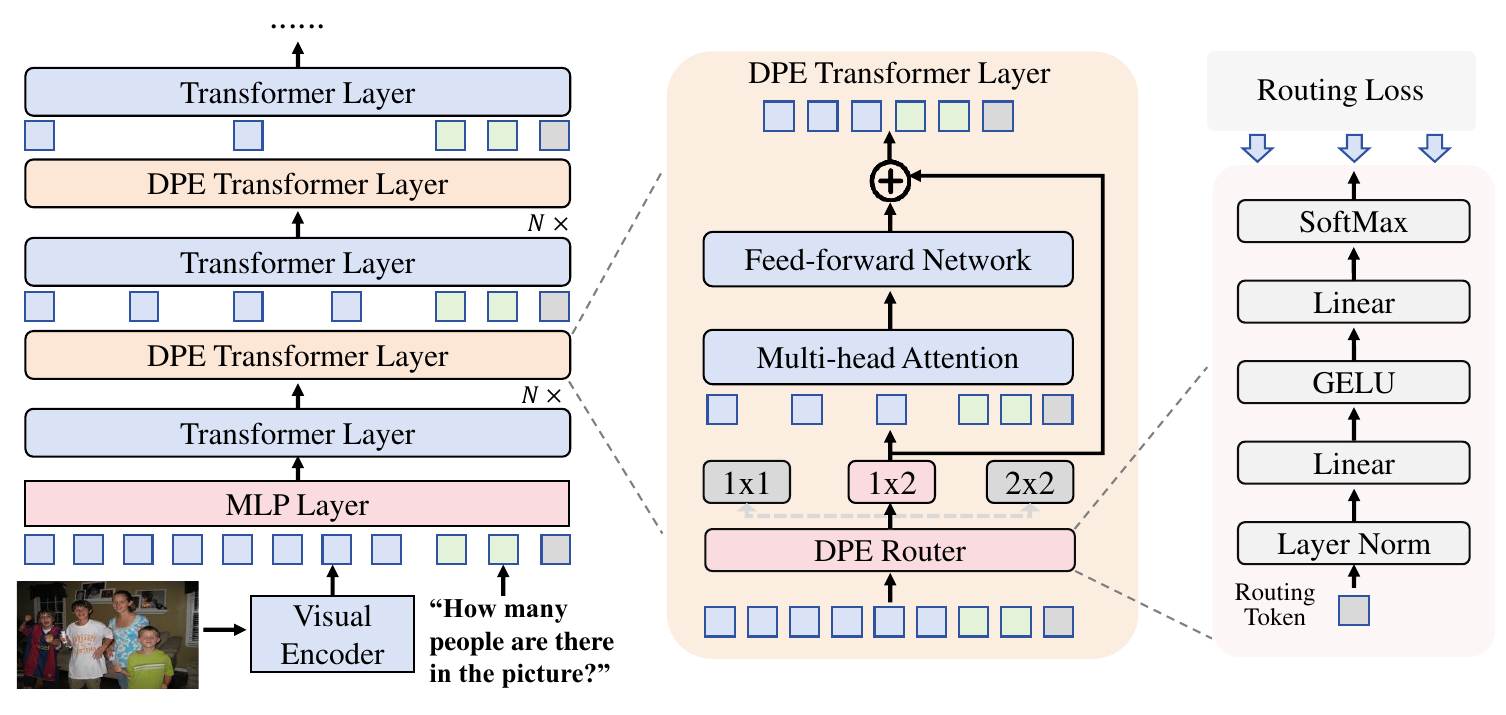}
   \caption{\textbf{Illustration of Dynamic Pyramid Network (DPN) and  its  Dynamic Pooling Experts (DPE).}  DPN formulates the common LLM as a dynamic pyramid structure, and the visual tokens will be progressively pooled via the DPE.  In practice, DPE can dynamically select an optimal pooling kernel for visual compression, thus achieving the best trade-off between efficiency and performance.  }
   \label{fig:framework}
\end{figure*}

Based on the above principle, we propose a novel dynamic pyramid network for MLLMs, namely (DPN). As shown in Fig.~\ref{fig:framework},  DPN not only adopts a pyramid structure to realize effective  visual compression, but can also dynamically adjust the compression rate based on samples.  To approach this target, DPN is equipped with a set of dynamic pooling experts (DPE), which aim to select an optimal pooling rate based on input features.  To maximize the benefits of DPE, we propose a novel routing loss to encourage visual sparsity while preserving performance.  With these designs, DPN can be seamlessly deployed on existing MLLMs without additional training stages.

To validate our approach, we adopt DPN to two popular MLLMs, \emph{i.e.,} LLaVA~\cite{liu2023visualinstructiontuning} and LLaVA-HR~\cite{luo2024feasteyesmixtureofresolutionadaptation}, and conduct extensive experiments on 10 MLLM benchmarks, \emph{e.g.,} MM-Vet~\cite{yu2023mmvetevaluatinglargemultimodal}. Experiments not only confirm the great efficiency of DPN against existing methods,  but also validate its competitive performance compared to the baseline.  For example, compared to LLaVA-HR-X~\cite{luo2024feasteyesmixtureofresolutionadaptation}, our DPN-LLaVA-HR-X achieves 1.4$\times$ speedup and +0.62\% performance gains. In conclusion, our contributions can be summarized in three folds:

\begin{itemize} 
  \item We propose a novel Dynamic Pyramid Network (DPN) for multimodal large language models (MLLMs), which can achieve significant inference acceleration with minor performance sacrifice.
    \item  In DPN,  we propose an innovative Dynamic Pooling Experts (DPE) that can dynamically choose the optimal visual pooling rate based on input features. DPE is also equipped with a routing loss to maximize its benefits during training. 
    \item Compared to existing methods, DPN reduces up to 56\% average Flops on  10 benchmarks, while also well preserving the model performance, \emph{e.g.,} +0.74 gains on LLaVA. 
\end{itemize}

\section{Related Work}

\subsection{Pyramid Network}

Pyramid-based networks have demonstrated exceptional performance across various traditional tasks, including detection and segmentation. One such network, the Feature Pyramid Network (FPN)~\cite{lin2017featurepyramidnetworksobject}, leverages the inherent multi-scale, pyramidal hierarchy of deep convolutional networks to construct feature pyramids at a marginal additional cost. The FPN architecture utilizes a top-down approach with lateral connections to produce high-level semantic feature maps across multiple scales, which significantly enhances its utility as a generic feature extractor in numerous applications.
The concept of this pyramid-shaped network has been carried forward into the era of transformers\cite{vaswani2023attentionneed}. The Pyramid Vision Transformer (PVT)\cite{wang2021pyramidvisiontransformerversatile} integrates a progressive pyramid structure to reduce the computational burden of processing large feature maps. By combining the strengths of both convolutional neural networks (CNNs)\cite{o2015introduction} and transformers, PVT has achieved superior performance in downstream tasks such as object detection and instance segmentation.

% Building on this foundation, we propose incorporating the concept of pyramid networks into multimodal large language models. Based on the multi-step inference mechanism of the large language model, this pyramid-based approach could offer significant improvements in inference speed. Our proposed Dynamic Pooling Experts (DPE) layer introduces a dynamic pyramid structure optimized for large language models, balancing performance with inference acceleration. This design ultimately provides a more efficient inference mechanism, leading to enhanced model outcomes.

\subsection{Multimodal Large Language Models}

With the rise of large language models (LLMs)\cite{touvron2023llamaopenefficientfoundation,jiang2024mixtralexperts,bai2023qwentechnicalreport,openai2024gpt4technicalreport,vicuna2023,yang2024qwen2technicalreport}, an increasing number of researchers have turned their attention to the construction of multimodal large language models. The primary approach for extending LLMs into multimodal large language models (MLLMs)~\cite{lin2024vilapretrainingvisuallanguage,xu2024llavauhdlmmperceivingaspect,li2024llavaonevisioneasyvisualtask,liu2023visualinstructiontuning,luo2024feasteyesmixtureofresolutionadaptation,li2024minigeminiminingpotentialmultimodality,dai2023instructblipgeneralpurposevisionlanguagemodels,chu2023mobilevlmfaststrong,chu2024mobilevlmv2fasterstronger,wang2024qwen2vlenhancingvisionlanguagemodels} is connecting pre-trained image encoders, such as CLIP\cite{radford2021learningtransferablevisualmodels}, aligning image and text features, and fine-tuning on visual question answering tasks. For instance, early works like BLIP-2\cite{li2023blip2bootstrappinglanguageimagepretraining}  introduced Q-Former to efficiently bridge visual and language features. Models like LLaVA fully map pre-trained image features directly into the semantic space, achieving superior performance but at the cost of increased inference complexity. Additionally, some researchers have focused on the resolution limitations of pre-trained image encoders, developing high-resolution encoding strategies to further enhance the performance of multimodal large language models, as seen in works like LLaVA-HR. However, these high-resolution methods often increase the number of visual tokens, resulting in significantly higher computational costs. 

\subsection{Efficient Multimodal Large Language Models}

Recently, many studies have focused on accelerating multimodal large models, which can be broadly categorized into the exploration of efficient projectors ~\cite{chu2024mobilevlmv2fasterstronger,shang2024llavaprumergeadaptivetokenreduction,li2024tokenpackerefficientvisualprojector,lan2024avgllavalargemultimodalmodel,xiong2024pyraparallelyieldingreactivation} and the investigation of sparse structures~\cite{bolya2023tokenmergingvitfaster,shi2024crossgetcrossguidedensembletokens,cao2023pumerpruningmergingtokens,kong2022spvitenablingfastervision,liang2022patchesneedexpeditingvision} within LLMs ~\cite{chen2024imageworth12tokens,wu2024routingexpertslearningroute,xing2024pyramiddrop,luo2024gammamodexploringmixtureofdepthadaptation,lin2024moellavamixtureexpertslarge}. Taking the representative work in each direction as an example, Token Packer~\cite{li2024tokenpackerefficientvisualprojector} introduced an efficient projector that only needs to use one-quarter of the original visual tokens. MoE-LLaVA~\cite{lin2024moellavamixtureexpertslarge} explores sparse structures to reduce the number of activated parameters. FastV~\cite{chen2024imageworth12tokens} discards visual tokens in the LLM middle layer through the attention score. Routing Experts(RoE)~\cite{wu2024routingexpertslearningroute} allows some tokens to skip complex attention~\cite{vaswani2023attentionneed} layers and FFN layers through dynamic route prediction. Such methods often have obvious performance degradation in certain types of tasks, and the efficient projector method brings more training costs. Our proposed DPN performs dynamic compression within MLLMs, automatically selecting an appropriate compression ratio based on the complexity of the visual task. This approach maximizes inference efficiency while preserving the model's ability to perceive image details. In addition, our method provides a simpler training method, which can be integrated with the visual instruction tuning stage of MLLM and reuse the pre-trained projector, instead of requiring multi-stage training like RoE-LLaVA and MoE-LLaVA.

\section{Preliminaries}
We first recap the structure of existing MLLMs and their efficient variants.

\textbf{Plain MLLMs.} As shown in Fig.~\ref{fig1}, existing MLLMs typically adopt the ViT-MLP-LLM structure, \emph{e.g.,} LLaVA~\cite{liu2023visualinstructiontuning}. In particular, given the input image $I \in \mathbb{R}^{H \times  W \times 3}$, a ViT as visual encoder is used to extract the visual features, denoted by $X \in \mathbb{R}^{(h \times w) \times d}$.  After that, the visual features are projected into the embedding space of the LLM via an MLP layer. Finally, the visual features  $X \in \mathbb{R}^{(h \times w) \times d}$ are combined with the text features  $Y \in \mathbb{R}^{(t-1) \times d}$ and fed into the LLM to decode the words. Therefore, a typical MLLM can be formulated by:
\begin{equation}
    \begin{aligned}
        p_t= \mathcal{F_\text{llm}}(y_{t}|\mathcal{F_\text{proj}}( x; \theta_p), y_{0:t-1};\theta).
    \end{aligned} 
    \label{eq_1}
\end{equation}
Where $y \in Y$   and $p_t\in \mathbb{R}^{m}$ denote the text features   and the  next-token probability, respectively.  $\mathcal{F}_\text{llm}$  and  $\mathcal{F}_\text{proj}$ denote the LLM and the projector, respectively.  $\theta$ and $\theta_p$ are their corresponding parameters.   In Eq.~\ref{eq_1}, we can see that the computation cost of the MLLM is highly dependent on the number of the input features, \emph{i.e.,} $[F_v, F_t]$. In practice, the number of visual features is much longer than that of textual features, \emph{e.g.,} 576  \textit{vs.} $\sim$30 in LLaVA~\cite{liu2023visualinstructiontuning}.   

\textbf{Efficient Projector.} To maintain the model efficiency, numerous works  propose  the efficient projector to reduce the number of visual features~\cite{chu2024mobilevlmv2fasterstronger,shang2024llavaprumergeadaptivetokenreduction,li2024tokenpackerefficientvisualprojector,lan2024avgllavalargemultimodalmodel}.   In particular,  given a down-sampling rate $k$, existing efficient projectors aim to pool the visual features into a smaller shape. In this case, the MLLM can be re-formulated by:
\begin{equation}
        p_t= \mathcal{F_\text{llm}}(y_{t}|\mathcal{F_\text{proj}}( x; \theta_p,k), y_{0:t-1};\theta).
    \label{eq2}
\end{equation}
As shown in Eq.~\ref{eq2}, the  projector is placed outside the LLM and its visual compression  completely relies on the visual saliency or the pre-defined kernel. Such practices often lead to  two potential limitations: 1). Useful visual details will be eliminated before feeding into the LLM. 2).  A static compression rate struggles to  be optimal for all samples.

\section{Dynamic Pyramid Network}

\subsection{Overview}
To overcome the above limitations, we formulate the MLLM as a dynamic pyramid network (DPN). As shown in Fig.~\ref{fig:framework}, the most unique property of DPN lies in that the visual compression module is embedded into the LLM and the compression rate is also dynamically dependent on the input features.  Specifically,  a  DPN  layer can be defined by
\begin{equation}
\begin{aligned}
Z&=[\mathcal{F_\text{dpe}}( x; \theta_d,k_0, k_1, k_2), y],\\
Z&=\text{MHA}(\text{Norm}(Z))+Z,\\
Z&=\text{FFN}(\text{Norm}(Z))+Z'.
    \label{eq3}
\end{aligned}
\end{equation} 
Here, MHA, FFN, and Norm denote the multi-head attention~\cite{vaswani2023attentionneed}, the feed-forward network~\cite{vaswani2023attentionneed} and the normalization~\cite{touvron2023llamaopenefficientfoundation} 
, respectively. $[\cdot]$ is the concatenation operation. $\mathcal{F}_{\text{dpe}}$  and $\theta_d$  denote the dynamic pooling expert layer, which will dynamically select a pooling kernel from $\{k_0, k_1, k_2\}$ to down-sample the visual features.  As shown in Fig.~\ref{fig:framework}, we will replace the 8-\textit{th}, 16-\textit{th} and 24-\textit{th} layers of LLM with our DPN layer, yielding a novel dynamic pyramid structure.  Compared with previous studies, such a structure not only prevents fine-grained visual contents in shallow layers, but also maximizes efficiency according to the input samples.

\subsection{Dynamic Pooling Experts}
The key of Dynamic Pooling Experts (DPE) is to select an optimal visual compression rate based on the given samples, \emph{i.e.,} the visual and textual content.  Therefore, we formulated DPE as an mixture-of-expert layer, where each expert is defined as a projection layer with different compression rates.  To approach this target, we first insert a learnable routing token $r \in \mathbb{R}^{d}$ into the input features, \emph{i.e.,} $[x,y,r]$.  In this case, the routing token can aggregate the vision-language information of input features through the attention mechanism, thereby making the optimal decision for its routing. In particular, DPE is defined as:
\begin{equation}
      \mathcal{F}_{\text{dpe}}(x,r)=R(r)_iE_i(x),
\end{equation}
where $i=\text{Top}_{1}(R(r))$ indicates the index with the highest score. $R(\cdot)$ and $E(\cdot)$ denote the router and the pooling expert.   In our experiments, the pooling experts can be simply defined as the max pooling layers with a set of pre-defined kernels $\{k_0, k_1, k_2\}$.  And the router $R(\cdot)$ can be written as:
\begin{equation}
      R(r)=\sigma(\text{MLP}(r)).
      \label{router}
\end{equation}
Here, $\sigma(\cdot)$ is the softmax function.  As defined in Eq.~\ref{router},  the router will predict scores to indicate the significance for each pooling expert.   
During inference, DPE will  select the pooling expert with the highest probability  to  reduce  the number of visual tokens.

\begin{figure*}[t]
  \centering
   \includegraphics[width=1.0\linewidth]{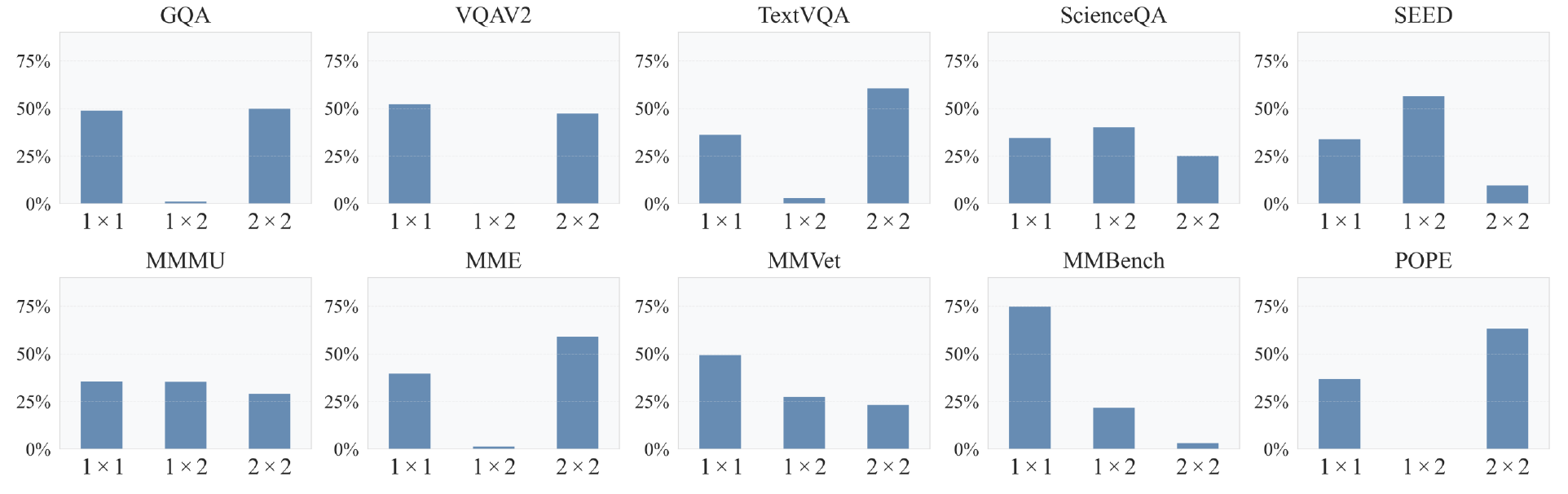}
   \caption{\textbf{Statistics of expert activation for different datasets. } our DPN can dynamically select the pooling kernel according to the task difficulty.}
   \label{fig: Datasets Routing Distribution}
\end{figure*}

\subsection{Routing Loss} 
Routing loss aims to maximize the visual sparsity while encouraging the routing diversity.  Without the routing loss,  the router will be completely optimized by the autoregressive loss and tends to select a constant expert. To this end, we design the routing loss as a hinge function that can keep the expected compression rate close to our pre-defined target. In this case, the routing loss can be written as:
\begin{equation}
  \mathcal{L}_{r} = \max(0,t- \frac{1}{n} \Sigma_{i=1}^n R(r)_i C_i ),
\end{equation}
where $C_i$ denotes the compression rate  of the i-\textit{th} pooling expert, \emph{e.g.,} 2 for 2$\times$ 2 pooling. $t$ is the target compression rate, which is set to 1.5 in our experiments. 
 
% We found that even without any additional loss, optimizing the probability coefficient of the vision token after pooling still has slightly diversified routing results based on the original autoregressive loss. However, most elements still tend to be assigned to a single expert. To achieve better acceleration, we draw inspiration from hinge loss and design a Pooling Reward Loss to increase routing diversity and encourage the selection of a larger kernel size.

% We set the reward coefficients for experts $E_1$, $E_2$, and $E_3$ to $s^{(1)}=0$, $s^{(2)}=1$, and $s^{(3)}=2$.

% We directly optimize the maximum expert probability of router prediction during training. In addition, we set a reward threshold and do not give excessive rewards, which will cause verification to affect the performance of the model. For multiple layers of routers, we optimize them separately and finally average the loss.
% Let $n$ be the number of samples, $t$ be the reward threshold, $p^{(i)}$ be the maximum predicted probability of the i-th sample, and $s^{(i)}$ be the reward coefficient of the expert corresponding to the maximum probability of the i-th sample. and the pooling reward loss can be expressed as
% \begin{equation}
%   L_{reward} = max(0,t- \frac{1}{n} \Sigma_{i=1}^n s^{(i)} p^{(i)} )
% \end{equation}

Therefore, the overall learning objective of the MLLM is
\begin{equation}
  \mathcal{L}= \mathcal{L}_{a} + \lambda \cdot \mathcal{L}_{r}. 
\end{equation}
Here, $\mathcal{L}_{a}$ denotes the original autoregressive loss of the MLLM. $\lambda$ is a coefficient to balance the contribution of two losses, which is set to 0.01 in our experiments.

\subsection{The Deployment on MLLMs}
Our DPN can be seamlessly deployed on most existing MLLMs, \emph{e.g., } LLaVA~\cite{liu2023visualinstructiontuning} and LLaVA-HR~\cite{luo2024feasteyesmixtureofresolutionadaptation}.  In existing MLLMs, training is typically divided into two stages, \emph{i.e.,} vision-text alignment and visual instruction tuning. Our DPN is deployed in the instruction tuning stage, and replaces  8-\textit{th}, 16-\textit{th} and 24-\textit{th} standard Transformer layers of the MLLM with our DPE layers. Therefore, we can directly skip the alignment stage by using existing pre-trained weights.  During inference, our DPN can directly conduct dynamic inference without any modification.

\subsection{Efficiency Analysis}
For FLOPs estimation, we follow the calculation method of FastV\cite{chen2024imageworth12tokens} and only consider the multi-head attention (MHA) layer and feed-forward network (FFN) layer. For one transformer layer, assume $n$ is the token number, $d$ is the hidden state size, $m$ is the intermediate size of FFN, the total FLOPs can be estimated by $4nd^2+2n^2d+2ndm$. Our method will lead to different number of tokens in  different layers. Assume $n_i$  is the number of tokens in the i-\textit{th} layer,   the theoretical FLOPs ratio of our method compared to the baseline is computed as:
\begin{equation}
\frac{\Sigma_{i=1}^{I} 4n_{i}d^2+2n_i^2d+2n_idm}{I\times (4nd^2+2n^2d+2ndm)}.
\end{equation}
For the overall FLOPs of our method, we average the single FLOPs value of all samples. In practice, $n_i << n$  and the overall FLOPs will be greatly reduced via our DPN.

% \begin{table*}[ht]
%   \centering
%   \begin{tabular}{l|c|c|c|c|c|c|c|c|c|c|c|c}
%     \toprule
%       &GQA & SQA\textsuperscript{I} & VQA\textsuperscript{T} & VQA\textsuperscript{v2} &MME & MM-Vet & MMB & MMMU & POPE & SEED& Average \\
%     DPE-LLaVA & $39\%$ & $45\%$ & $36\%$ & $39\%$ & $36\%$ & $45\%$ & $75\%$& $44\%$ & $35\%$ & $47\%$& $44\%$\\
%      % DPE-LLaVA-HR & $64\%$ & $57\%$ & $58\%$ & $62\%$ & $55\%$ & $58\%$ & $64\%$& $56\%$ & $59\%$\\ 
%     \bottomrule
%   \end{tabular}
%   \caption{Acceleration Details: This table shows the speedup details of DPE on different datasets.}
%   \label{tab: Acceleration Details}
% \end{table*}

\begin{table*}[ht]
  \centering
  \scalebox{0.76}{
  \begin{tabular}{lc|cccccccccc|ccc}
    \toprule
    Methods & Params& GQA & SQA\textsuperscript{I} & VQA\textsuperscript{T} & VQA\textsuperscript{v2} &MME & MM-Vet & MMB & MMMU & SEED\textsuperscript{I} & POPE & AVG & FLOPs  & Speed  \\
    \midrule
    I-80B \cite{laurençon2023obelicsopenwebscalefiltered} & 65B & 45.2 & - & -& 60.0 & -&-&54.5&-&-&-&-&-&- \\
    InstructBLIP \cite{dai2023instructblipgeneralpurposevisionlanguagemodels} & 14B & 49.5 & 63.1 & 50.7 & - & 1212 & 25.6 & - & - & - & 78.9 & - & - & - \\
    VILA \cite{lin2024vilapretrainingvisuallanguage}& 7B & 62.3 & 68.2 & 64.4 & 79.9& 1533 & 34.9 & 68.9 & - & - & 85.5 & - & - & -\\
    Qwen-VL\cite{bai2023qwenvlversatilevisionlanguagemodel} & 10B & 59.3 & 67.1 & 63.8 & 78.8 &  1487 & - & 38.2 & - &  - & - & - & - & 5.3\\
     \midrule
    MoE-LLaVA \cite{laurençon2023obelicsopenwebscalefiltered} & 2.7B×4  & 62.5 & 67.4 & 64.6 & 51.4 &1423&  34.3 & 65.2& -& - & 86.3 & 62.9 & -& 6.1\\%  68.0
    \midrule
    LLaVA 1.5 \cite{liu2023visualinstructiontuning} & 7B & 62.0 & 66.8 & 58.2 & 78.5 & 1510  & 31.1 & 62.1& 35.3& 66.2& 85.9 & 65.0 & 2986 & 9.3\\
    % TokenPacker & 7B & 61.8 & 70.6 &  57.16 & 77.9 & 1450 & 33.0 & 65.1 & 33.5 & 65.0 & 87.0\\
    RoE-LLaVA \cite{wu2024routingexpertslearningroute} &7B& 61.4 & 68.4 & 56.8 & \textbf{80.3} & \textbf{1522}& 31.9& 64.3 & - & -
    &86.1 & 65.2 & 2039 & 9.8\\ % 66.752
    % PruMerge+ & 7B & - & 68.3& 57.1 & 76.8 &1462 & & 64.9 &- &- & 84.0\\
    FastV(K=3,R=0.5) \cite{chen2024imageworth12tokens}& 7B & 60.4 & \textbf{68.6} & 45.5 & 75.8 & 1510& \textbf{32.4} & 63.8 & 35.2 & 
    65.3 & 85.0 &63.4 & 1431 & 10.8\\ 
    FastV(K=3,R=0.6) \cite{chen2024imageworth12tokens} &7B& 59.1 & 68.5 & 44.7 & 75.1 & 1496& 30.2 &63.8 & 36.3 & 
     64.4 &83.7 & 62.5 & 1192 & 12.3\\
    \textbf{DPN-LLaVA}&7B & \textbf{63.2} & 67.7 & \textbf{57.9} & 79.3 & 1469 & 28.7 &\textbf{65.8} & \textbf{36.8} & \textbf{68.4} & \textbf{87.7} & \textbf{65.5} & \textbf{1137} & \textbf{12.7}\\
   
    \midrule
    LLaVA-HR \cite{luo2024feasteyesmixtureofresolutionadaptation} & 7B & 64.2   & 67.9 & 67.1 & 81.9 & 1547 & 31.5 & 65.0 & \textbf{35.3} & 64.2 & 87.6 & 67.8 & 5429 & 5.0\\ % 70.2
    RoE-LLaVA-HR \cite{wu2024routingexpertslearningroute} & 7B & 62.5 & 67.4 & 64.6 & 80.9  & \textbf{1558} & 30.0 & 64.6 & -  & -& 88.1 & 67.0 & 4940 & 5.0 \\ % 68.85
    \textbf{DPN-LLaVA-HR}&7B  & \textbf{63.9}  & \textbf{69.3} & \textbf{65.8} & \textbf{81.5} & 1471 & \textbf{32.2} & \textbf{64.8} & 34.6 & \textbf{70.4} &\textbf{88.2} & \textbf{67.4} & \textbf{3257} & \textbf{7.1}\\ %  70.1
    \midrule
    LLaVA-HR-X  \cite{luo2024feasteyesmixtureofresolutionadaptation} & 13B & 65.2 & 69.7 & \textbf{70.9} & 82.9& \textbf{1487} & \textbf{40.3} &64.5 & \textbf{36.6} & 71.7& 88.8 & 69.4 & 10522 & 2.8\\
    \textbf{DPN-LLaVA-HR-X} & 13B & 64.6 & \textbf{71.0} & 70.1 & \textbf{82.9} & 1473 & 38.4 &\textbf{66.3} &34.4 &\textbf{71.7} &\textbf{90.9} & \textbf{69.7} & \textbf{5892} & \textbf{4.1}\\
    \bottomrule
  \end{tabular}
  }
  \caption{\textbf{Comparison with existing efficient MLLMs.} ``AVG'' denotes the average performance of eight public datasets.  The best performance is highlighted in bold.  The speed is the average number of test samples  per second and FLOPs is measured in billions. }
  \label{tab: Comparison with existing acceleration methods}
\end{table*}

\begin{table*}[ht]
\centering
\begin{tabular}{lcccccccccc}
    \toprule
    \textbf{Methods } & \textbf{\#Token}   & \textbf{VQA\textsuperscript{v2}} & \textbf{GQA} & \textbf{POPE} & \textbf{VizWiz} & \textbf{VQA\textsuperscript{T}} & \textbf{MMMU} & \textbf{MMB}& \textbf{MME}& \textbf{Avg.} \\
    \midrule
    MLP & 576    & 78.5 & 62.0 & 85.9 & 50.0 & 58.20 & 35.2 & 64.3& 1510& 63.43 \\
    Average-Pooling & 144    & 76.2 & 61.6 & 86.8 & 48.0 & 56.55&34.7 & 64.3& 1443& 62.34 \\
    Resampler\cite{bai2023qwenvlversatilevisionlanguagemodel} & 144   &  75.1 & 58.4 & 84.7 & 51.9 & -&-& 63.1& -& - \\
    C-Abstractor\cite{cha2024honeybeelocalityenhancedprojectormultimodal} & 144   &  74.6 & 59.2 & 84.6 & 49.2 & -&-& 63.1& -& - \\
    Pixel-Shuffle \cite{chen2024fargpt4vclosinggap} & 144   & 76.2 & 60.1 & 85.9 & 48.8 & -&-& 64.0& -& - \\
    LDP-v2 \cite{chu2024mobilevlmv2fasterstronger} & 144    & 77.3 & 61.1 & 86.1 & 47.6 & -&-& 66.2& - & - \\
    PruMerge +\cite{shang2024llavaprumergeadaptivetokenreduction}& \textasciitilde 144 & 76.8 & - & 84.0 & -& 57.10& -& 60.9 & 1462 & -\\ 
    TokenPacker \cite{li2024tokenpackerefficientvisualprojector} & 144    & 77.9 & 61.8 & 87.0 & \textbf{52.0} & 57.16 &35.9 & 65.1& 1450& 63.68 \\
    \midrule
    Ours & \textasciitilde 232 & \textbf{79.3}& \textbf{63.2} & \textbf{87.7} & 51.7& \textbf{57.93} & \textbf{36.8}&  \textbf{65.8}& \textbf{1469}& \textbf{64.48}\\
    \bottomrule
\end{tabular}
\caption{\textbf{Comparison with  existing projector-based acceleration methods.} Replacing the projector requires more training cost, while our method is simpler and can achieve more stable performance.}
\label{tab: Comparison with other efficient projector acceleration methods}
\end{table*}

\section{Experiments}

\subsection{Datasets and Metrics}
We test DPN on ten extensive public datasets, including four common vision-language benchmarks, including VQAv2~\citep{goyal2017makingvvqamatter}, GQA~\citep{hudson2019gqanewdatasetrealworld}, ScienceQA~\citep{lu2022learnexplainmultimodalreasoning}, and TextVQA~\citep{singh2019vqamodelsread} that evaluate basic object recognition, spatial reasoning, and text reading in images. We report results in their default settings. In addition, we evaluate DPN on six multimodal benchmarks for MLLMs, including MME~\citep{fu2024mmecomprehensiveevaluationbenchmark}, MMB~\citep{liu2024mmbenchmultimodalmodelallaround}, MM-Vet~\citep{yu2023mmvetevaluatinglargemultimodal}, MMMU~\citep{yue2023mmmu}, SEED~\citep{li2023seed}, POPE~\citep{Li-hallucination-2023} that assess broader multimodal comprehension and abstract reasoning. We report all the results in their default settings. For MME, we report the perception score.
\subsection{Implementation Details}
We validate our method on LLaVA~\citep{liu2023visualinstructiontuning} and LLaVA-HR~\citep{luo2024feasteyesmixtureofresolutionadaptation} followed by their default settings and reuse their pre-trained image projector from the visual-language alignment stage. In the visual instruction tuning stage, finetuning is conducted on LLaVA's 665K vision instruction dataset~\citep{liu2023visualinstructiontuning}; we insert our Dynamic Pooling Experts layer and train it simultaneously. For the 7B model with 32 transformer layers, we insert DPE after the 8th, 16th, and 24th layers. For the 13B model with 40 transformer layers, we insert DPE after the 10th, 20th, and 30th layers. During fine-tuning, the routing loss ratio is set to 0.01 for LLaVA and 1.0 for LLaVA-HR, depending on the input's resolution.

\begin{table*}[ht]
  \centering
  \begin{tabular}{lc|cccccc|c}
    \toprule
    Methods & Layers & GQA  & VQA\textsuperscript{T}  & SEED \textsuperscript{I} & POPE & MMMU & MMB & Average \\
    \midrule
     static $2\times 1$ pooling & 8,16,24& 62.34 & 57.39  & 66.6& 87.1 &34.7  & 64.4 & 62.08\\
    static $1\times 2$ pooling & 8,16,24& 62.12 & 57.70  & 67.43& 87.8 & 34.6 & 63.6 & 62.20\\
    static $2\times 2$ pooling & 8,16,24& 63.03 & 57.57 & 67.43 & 87.7 & 33.4 & 65.0 & 62.36\\
    dynamic  $2\times 2$,$4\times 4$ pooling & 8,16,24 & 60.72& 54.94 & 62.33 & 86.4 & 34.6 & 59.3 & 59.72\\
    dynamic  $1\times 2$,$2\times 2$ pooling & 4,8,12 & 59.86 & 53.92 & 63.02 & 85.7 & 34.4 & 60.2 & 59.52\\
    dynamic $1\times 2$,$2\times 2$ pooling & 8,16,24 & \textbf{63.17} & \textbf{57.93} & \textbf{68.36} & 87.7 & \textbf{36.8} & \textbf{65.8} & \textbf{63.29}\\
    \bottomrule
  \end{tabular}
  \caption{\textbf{Comparison static pyramid network with our DPN.} ``Static'' denotes that the routing scheme is not used and visual tokens are directly pooled by a constant kernel. ``Layers'' denotes the placement of the DPE. }
  \label{tab: dynamic pooling experts ablation study}
\end{table*}

\begin{table*}[ht]
  \centering
  \scalebox{1.0}{
  \begin{tabular}{l|cc|cc|cc|cc|cc}
    \toprule
    Methods  &  \multicolumn{2}{c|}{GQA}  & \multicolumn{2}{c|}{VQA\textsuperscript{T}} & \multicolumn{2}{c|}{POPE}& \multicolumn{2}{c|}{MMMU}&  \multicolumn{2}{c}{Average}
    \\
     & Metrics & FR & Metrics & FR& Metrics & FR& Metrics & FR & Metrics & FR\\
    \midrule
     Baseline & 62 & 100\% & 58.2 & 100\% & 85.9 & 100\% & 35.2 & 100\% & 60.3 & 100\%\\
     + DPE  & 63.17 & 50\% & 58.4 & 47\% & 87.9 &  48.4\% & 35.1 & 40\% & 61.1 & 46\% \\ 
     \textbf{+ Routing Loss} &\textbf{63.17}  &\textbf{39\%} & 57.93 & \textbf{36\%} & 87.7 & \textbf{35\%} & \textbf{36.8} & 44\% & \textbf{61.4} & \textbf{38.5\%}\\
    \bottomrule
  \end{tabular}
  }
  \caption{\textbf{Ablation Study of DPN on LLaVA.} Our baseline is the original LLaVA without any modification. ``FR'' denotes the FLOPs ratio.}
  \label{tab: Pooling Reward Loss Ablation Study}
\end{table*}

\subsection{Experimental Results}
\subsubsection{Ablation Study} 

We conduct a series of ablation experiments in Table~\ref{tab: dynamic pooling experts ablation study} and Table~\ref{tab: Pooling Reward Loss Ablation Study} to validate each design in our DPN framework. 

First, we examine naive pooling strategies (e.g., pre-pooling or random pooling), which fail on certain benchmarks and substantially degrade performance, confirming that simplistic pooling schedules cannot effectively preserve accuracy. In contrast, our \emph{Dynamic Pooling Experts} (DPE) mechanism adaptively selects the most suitable pooling kernel size at deeper layers, improving GQA by +2.27\% and VQA\textsuperscript{T} by +1.2\% over the base model.

We explored the selection of pooling kernel sizes in the first, second, and third rows of Table~\ref{tab: dynamic pooling experts ablation study}. Using $1\times2$ max pooling to merge adjacent tokens is an intuitive approach that better aligns with the autoregressive nature of the model. We also experimented with the $2\times1$ pooling approach, but its overall performance was inferior to that of $1\times2$ pooling. Since both methods achieve the same acceleration ratio, our final approach adopts dynamic $1\times2$ and $2\times2$ pooling. 

We also explored larger $4\times4$ pooling in the 4-th and 5-th rows of Table \ref{tab: dynamic pooling experts ablation study}, as well as the insertion of DPE at earlier layers (4th, 8th, and 12th). These modifications slightly impacted performance. We believe that introducing excessive compression during training increases the learning difficulty for the model. Therefore, inserting pooling layers evenly across different model depths and adopting dynamic $1\times2$ and $2\times2$ pooling provides better generalization and performance.

Moreover, when we further incorporate the \emph{Routing Loss}, the FLOPs ratio is reduced from 50\% to 39\% on GQA, and metrics on MMMU increase by +1.7\%. As shown in Table~\ref{tab: dynamic pooling experts ablation study}, this approach not only stabilizes training but also achieves significant efficiency gains on tasks, where we observe up to -13.4\% FLOPs reduction compared to the baseline. Notably, our final design (Dynamic $1\times2$, $2\times2$ Pooling) achieves the highest average performance of 63.29\%, surpassing the static $2\times2$ pooling baseline by +1.21\%. These results clearly demonstrate the effectiveness of each proposed component in achieving an excellent trade-off between efficiency and performance.

\subsection{Comparison with Existing Methods} 

As shown in Table~\ref{tab: Comparison with existing acceleration methods}, we provide a detailed comparison of MLLM deployed by DPN with both dense and sparse models on 10 benchmarks. DPN consistently achieves superior performance across multiple benchmarks while performing as the most efficient method. For example, compared to LLaVA 1.5, DPN-LLaVA (7B) achieves a +1.2\% improvement on the GQA benchmark and a +2.2\% gain on the POPE metric. Moreover, our approach reduces the required FLOPs by over 50\% and boosts the inference speed by more than 36\%, clearly demonstrating a better trade-off between performance and efficiency.
Table~\ref{tab: Comparison with other efficient projector acceleration methods} further illustrates the advantages of our method when compared to various efficient projector acceleration techniques, such as MLP, Average-Pooling, and TokenPacker. With approximately 232 tokens, our approach achieves the highest scores in key tasks—for instance, on the VQA\textsuperscript{v2} benchmark, our method records a score of 79.3\%, which is +0.8\% higher than the closest competitor MLP, and on the GQA task it outperforms MLP methods by around +1.2\% while having around 59.7\% fewer tokens. These improvements in performance and efficiency, along with  simplicity of our training process, further confirm  superior performance and efficiency of DPN.

\begin{figure*}[ht]
  \centering
   \includegraphics[width=0.93\linewidth]{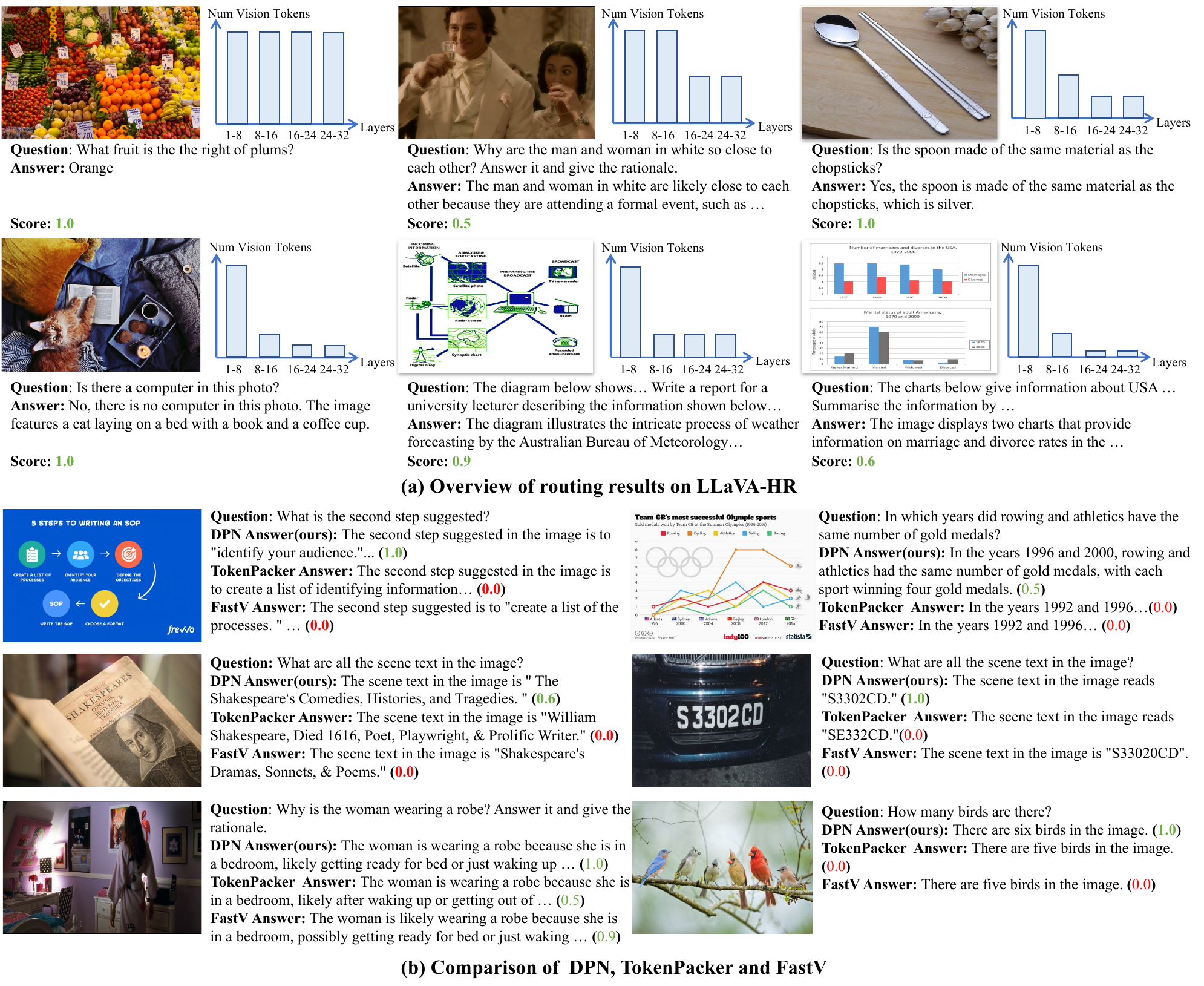}
   \caption{\textbf{Visualization Results.} Sub-figure A illustrates the routing of the DPN method across different visual tasks. Sub-figure B presents a visual comparison with the sparsification method FastV and the efficient projector-based approach TokenPacker. Our method demonstrates significant advantages in tasks such as optical character recognition (OCR), chart interpretation, and object detection.}
   \label{fig: Visualization Results}
\end{figure*}

In Fig.~\ref{fig: Visualization Results}, we further perform a visual analysis comparison with relevant methods to demonstrate that our approach can adjust visual token allocation based on image complexity and task specificity, highlighting its superiority in fine-grained understanding tasks. For instance, in the chart comprehension task shown in the first row of Fig.~b, existing methods exhibit noticeable errors. When asked, ``What is the second step suggested?", both Fast V and TokenPacker provided incorrect answers, stating ``create list of the processes", whereas our model correctly understood and captured the right response: ``identify your audience". Similarly, in the text recognition task depicted in the second row, other methods also performed poorly, either missing or misidentifying certain characters, while our approach produced accurate results. Additionally, existing methods made errors in the counting task, as seen in the second image of the third row, where six birds were mistakenly identified as five. These errors highlight that direct visual compression or indiscriminate token removal can lead to significant inaccuracies. In contrast, the DPN we propose applies progressive dynamic compression in MLLMs, better preserving their ability to perceive image details.

\subsubsection{Quantitative Analysis}

\noindent\textbf{The Diversity of Routing.}
We conducted a statistical analysis of the expert selection process for DPN-LLaVA 1.5 7B across different datasets, as presented in Fig.~\ref{fig: Datasets Routing Distribution}. It is important to note that each sample undergoes expert selection three times. For simplicity, we aggregate these selections without distinguishing between layers. Generally, each dataset results in 4 to 7 distinct routing patterns.

\noindent\textbf{The Impact of Image Complexity on Routing.}
Next, we analyzed the routing tendencies of DPE across various VQA samples, sampling from different routing patterns and visualizing the results in Fig.~\ref{fig: Visualization Results} . Our observations reveal that for complex images, DPE tends to avoid pooling. For instance, in the first row, the first image features a variety of fruits and text, prompting minimal pooling. In contrast, for simpler images containing only one or a few primary objects, DPE typically applies a $1 \times 2$ pooling, reducing the visual tokens by $50\%$, as shown in the second picture and third picture in the first row. For even simpler images, such as charts, DPE often utilizes $2 \times 2$ pooling, reducing $75\%$ of the visual tokens, as illustrated in the fourth and fifth rows. This behavior is intuitive, as charts contain more structured and densely packed information, making them easier to represent with fewer tokens than photographs.

\noindent\textbf{The Impact of Question Complexity on Routing.}
Finally, we perform simpler visual tasks on the first and second images of Fig.~\ref{fig: Visualization Results} to explore the impact of visual tasks on routing. The results indicate that DPN can recognize simpler visual tasks and perform inference more efficiently.

% \begin{figure}[ht]
%   \centering
%    \includegraphics[width=1\linewidth]{Routing Results 3.pdf}
%    \caption{Visualize DPE Routing Results for Different Questions}
%    \label{fig: Visualize Routing Results for Different Questions}
% \end{figure}

For the complex fruit image in the first column, the original question likely prompted the model to focus on intricate details, such as identifying specific types of fruit or distinguishing between similarly colored objects. However, when we simplified the question to ``Is there any fruit?" the model no longer needed to process such fine-grained information. Instead, it relied on broader, more generalized visual features to determine the presence of fruit. In this case, the DPE model selected a larger kernel size, which allowed it to process larger regions of the image at once. This decision not only reduced the computational load but also improved processing speed, as larger kernels aggregate information more efficiently over wider areas.
In the case of the second image, the original reasoning-based question required extensive contextual information—such as the mood conveyed by the photo, the environment, and the clothing and expressions of the individuals. All of these factors influenced the model's reasoning process. We switched to a simpler counting task, ``How many main characters are there in the picture?", for testing. In this setup, DPN applies a 50\% compression in the first and second DPE layer, achieving greater efficiency.

\section{Conclusions}

Existing acceleration methods, such as sparsification structures in MLLMs or efficient projector-based approaches, often compromise the perceptual capabilities of MLLMs. In this paper, we propose a novel Dynamic Pyramid Network (DPN) designed for efficient MLLMs, in which visual representations undergo progressive dynamic compression as depth increases. Under this framework, even with a high compression ratio, the model can still capture fine-grained visual details at shallow layers.We introduced Dynamic Pooling Experts (DPE) and Routing Loss to further improve inference efficiency without affecting performance or adding extra training overhead. We verified the necessity of each module design through ablation studies. We conduct extensive experiments on two popular MLLMs and ten benchmarks. The results demonstrate that DPN reduces FLOPs by 56\% on LLaVA while achieving a 0.74\% performance improvement, outperforming existing methods.  Furthermore, the generalization capability of DPN is validated on the high-resolution MLLM LLaVA-HR. We also provide a visual analysis comparing expert routing with related approaches, showing that DPN dynamically selects the optimal visual compression ratio based on input features and shows obvious advantages in various VL tasks.

% \section{Final copy}

% You must include your signed IEEE copyright release form when you submit your finished paper.
% We MUST have this form before your paper can be published in the proceedings.

% Please direct any questions to the production editor in charge of these proceedings at the IEEE Computer Society Press:
% \url{https://www.computer.org/about/contact}.
{
    \small
    \bibliographystyle{ieeenat_fullname}
    \bibliography{main}
}

\end{document}